\documentclass[10pt,twocolumn,letterpaper]{article}

\usepackage[pagenumbers]{cvpr} 

\usepackage{graphicx}
\usepackage{subcaption}
\usepackage{enumitem} 
\usepackage{multirow} 
\usepackage{makecell}
\usepackage{pifont}
\usepackage{amssymb}
\usepackage{amsmath}
\usepackage{booktabs}
\usepackage{url}            

\definecolor{cvprblue}{rgb}{0.21,0.49,0.74}
\usepackage[pagebackref,breaklinks,colorlinks,allcolors=cvprblue]{hyperref}

\def\paperID{5696} 
\def\confName{CVPR}
\def\confYear{2025}

\title{Partial Channel Network: Compute Fewer, Perform Better}

\author{Haiduo Huang, Tian Xia, Wenzhe zhao, Pengju Ren \\
Xi'an Jiaotong University $\cdot$  Institute of Artificial Intelligence and Robotics\\
{\tt\small huanghd@stu.xjtu.edu.cn, stian\_xia@xjtu.edu.cn, wenzhe@xjtu.edu.cn, pengjuren@xjtu.edu.cn}
}

\begin{document}
\maketitle
\begin{abstract}
    Designing a module or mechanism that enables a network to maintain low parameters and FLOPs without sacrificing accuracy and throughput remains a challenge. To address this challenge and exploit the redundancy within feature map channels, we propose a new solution: {\bf p}artial {\bf c}hannel {\bf m}echanism ({\bf PCM}). Specifically, through the split operation, the feature map channels are divided into different parts, with each part corresponding to different operations, such as convolution, attention, pooling, and identity mapping. Based on this assumption, we introduce a novel {\bf p}artial {\bf at}tention {\bf conv}olution ({\bf PATConv}) that can efficiently combine convolution with visual attention. Our exploration indicates that the PATConv can completely replace both the regular convolution and the regular visual attention while reducing model parameters and FLOPs. Moreover, PATConv can derive three new types of blocks: Partial Channel-Attention block (PAT\_ch), Partial Spatial-Attention block (PAT\_sp) and Partial Self-Attention block (PAT\_sf). In addition, we propose a novel {\bf d}ynamic {\bf p}artial {\bf conv}olution ({\bf DPConv}) that can adaptively learn the proportion of split channels in different layers to achieve better trade-offs. Building on PATConv and DPConv, we propose a new hybrid network family, named {\bf PartialNet}, which achieves superior top-1 accuracy and inference speed compared to some SOTA models on ImageNet-1K classification and excels in both detection and segmentation on the COCO dataset. Our code is available at \url{https://github.com/haiduo/PartialNet}.
\end{abstract}    
\section{Introduction}
\label{sec:intro}
Designing an efficient and effective neural network has remained a prominent topic in computer vision research. To design an efficient network, many prior works adopt depthwise separable convolution (DWConv)~\cite{Howard2017} as a substitute for regular dense convolution. For instance, some CNN-based models~\cite{Sandler2018a, Tan2019} leverage DWConv to reduce the model's FLOPs and parameters, while Hybrid-based models~\cite{yang2022focal, hou2022conv2former, Rao2022} employ DWConv to simulate self-attention operations to decrease computation complexity. However, some studies~\cite{Ma2018, ding2022scaling} have revealed that DWConv may suffer from frequent memory access and low parallelism during inference~\cite{Chen2023}, which leads to low throughput.

\begin{figure}[ht]
  \centering
  \includegraphics[width=0.99\linewidth]{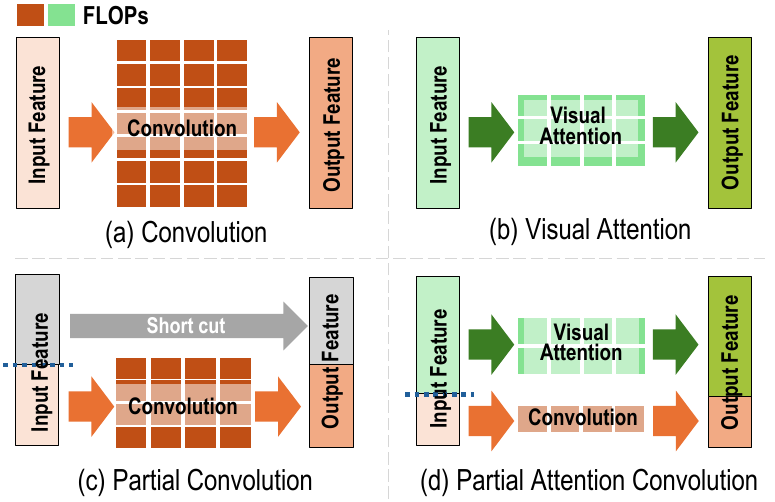}
  \caption{Comparison of different operation types.}
  \label{fig:diff_modul}
\end{figure}

Considering the substantial redundancy between feature maps~\cite{Han2020, Chen2023}, to further reduce the computational cost and parameters while improving the model's inference speed and accuracy, we introduce the partial channel mechanism (PCM) to fully exploit the value of feature map channels. PCM primarily splits the feature maps into different parts by a split operation, with each part undergoing different operations, followed by an integration by a concatenation operation. Our insight is that by reasonably allocating different operations and using cheaper, efficient operators to partially replace costly, dense operators, it is possible to improve the inference speed of a model while further enhancing its accuracy. Specifically, our approach involves replacing computationally intensive convolutions with computationally cheaper visual attention to enhance the model's representation capability and improve inference speed. Based on this idea, we introduce a novel Partial Attention Convolution (PATConv), which can replace both regular convolutions and regular visual attention while also reducing the model's parameters and FLOPs. This approach primarily integrates partial convolution with partial visual attention, as illustrated in~\cref{fig:diff_modul}.

Some recent works split the feature map, but they use a naive approach, where one part is operated on while the other part is left untouched. For instance, ShuffleNet V2~\cite{Ma2018} introduces a ``Channel Split" operation that divides the input feature channels into two parts: one part is retained directly, while the other part undergoes multiple layers of convolution. Similarly, FasterNet~\cite{Chen2023} introduces a partial convolution (PConv) that selectively applies Conv to a subset of input channels, leaving the remaining channels untouched. This makes PConv faster in inference speed compared to both regular Conv and regular DWConv. However, they merely consider improving the model's performance from the perspective of reducing the number of computed feature channels, neglecting the potential value of other channel portions. 

Therefore, we believe that it is more reasonable to account for the overall FLOPs, inference speed, and accuracy from a global perspective by exploiting the entire feature channels. Specifically, we propose to combine Conv and attention, applying them to partial channels respectively. Since there are no effective attention algorithms for partial channels, we invent three efficient partial visual attention blocks: (1) PAT\_ch integrates an enhanced Gaussian channel attention mechanism~\cite{Hu2018}, facilitating richer inter-channel global information interaction. (2) PAT\_sp introduces the spatial-wise attention to the MLP layer to further improve model accuracy. (3) PAT\_sf refers to the MetaFormer-based~\cite{Yu2022a} paradigm and integrate global self-attention into the last stage of the model to expand its global receptive field. 

Although the above improvements can provide a stronger visual model, considering the limitations of latency and parameters, the ratio of split channels between partial convolution and partial visual attention needs to be further optimal. To address this problem, we refer to dynamic group convolution~\cite{Zhang2019c} and propose a novel dynamic partial convolution, which can effectively and adaptively learn the split ratio of different layers according to the constraints (\eg, parameters and latency) to achieve the optimal trade-off between model inference speed and accuracy.

\begin{figure}[ht]
  \centering
  \includegraphics[width=1.\linewidth]{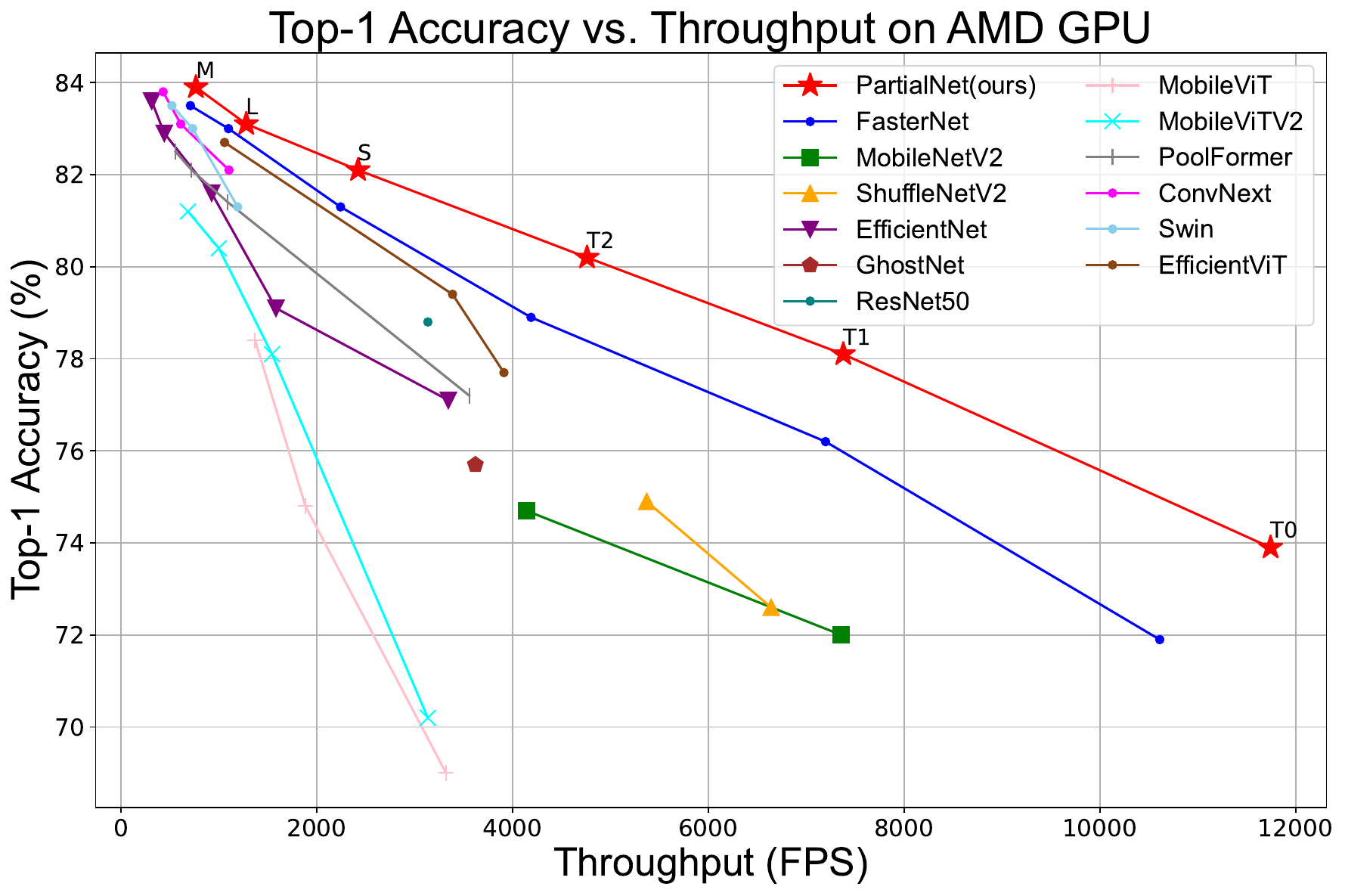}
  \caption{Our PartialNet achieves higher trade-off of accuracy and throughput on ImageNet-1K.}
  \label{fig:acc_throughput}
\end{figure}

In conclusion, the enhanced model is dubbed {\bf PartialNet}, which achieves overall performance improvement in the ImageNet1K classification task while maintaining similar throughput, as is presented in~\cref{fig:acc_throughput}. Our main contributions can be described as:
\begin{itemize}
  \item We propose a partial channel mechanism and introduce a partial attention convolution (PATConv) that integrates visual attention into partial convolution in a parallel way, which differs from the series way of previous works and can improve model performance while increasing inference speed.
  \item Based on the PATConv, we develop three partial visual attention blocks: PAT\_ch exhibits high potential as a replacement for regular convolution and DWConv, PAT\_sp can effectively reinforce MLP layers at minimal cost, while PAT\_sf integrates local and global features, achieving higher accuracy.
  \item To achieve a better trade-off between model inference speed and accuracy, we propose a novel dynamic partial convolution (DPConv), which can adaptively learn the split ratio of different layers according to the constraints (\eg, model parameters).
  \item Building upon the above methods, we design a new hybrid-based model family named PartialNet that shows improved performance on standard vision benchmarks over most efficient SOTA models.
\end{itemize}

\section{Related Work}
\label{sec:related_work}
~~~{\bf Efficient CNNs and ViTs.}
DWConv is widely adopted in the design of efficient neural networks, such as MobileNets~\cite{Sandler2018a, Howard2019}, EfficientNets~\cite{Tan2019, tan2021efficientnetv2}, MobileViT~\cite{Mehta2021}, and EdgeViT~\cite{Pan2022}. Because of its efficiency limitations on modern parallel devices, numerous works have aimed to improve it. For example, RepLKNet~\cite{ding2022scaling} uses larger-kernel DWConv to alleviate the issue of underutilized calculations. PoolFormer~\cite{Yu2022a} achieves strong performance through spatial interaction with pooling operations alone. Recently, FasterNet~\cite{Chen2023} reduces FLOPs and memory accesses simultaneously by introducing partial convolution. Nevertheless, FasterNet does not outperform other vision models in accuracy. In contrast, our proposed PartialNet addresses this limitation by integrating the visual attention into convolution to enhance the accuracy of models.

{\bf Visual Attention.}
The effectiveness of Vision Transformers (ViTs) mainly attributes their success to the role of attention mechanisms~\cite{raghu2021vision, paul2022vision}. In visual tasks, attention mechanisms are commonly categorized into three types: Channel Attention, Spatial Attention, and Self-Attention. Some works~\cite{Mehta2022, Rao2022, Shaker2023, Cai2023b} employ various techniques to implement the Self-Attention mechanism efficiently, \eg, Linear Attention~\cite{Wang2020c, Cai2023b}. Furthermore, the effectiveness of Channel Attention and Spatial Attention has already been validated in SRM~\cite{Lee2019a}, SE-Net~\cite{Hu2018} and CBAM~\cite{Woo2018}. Similarly, we have incorporated attention to the same feature, but in a parallel way with partial attention to mitigate the impact of element-wise multiplication on overall inference speed.

\section{Methodology}
\label{sec:method}
\begin{figure*}[ht]
  \centering
  \includegraphics[width=0.96\textwidth]{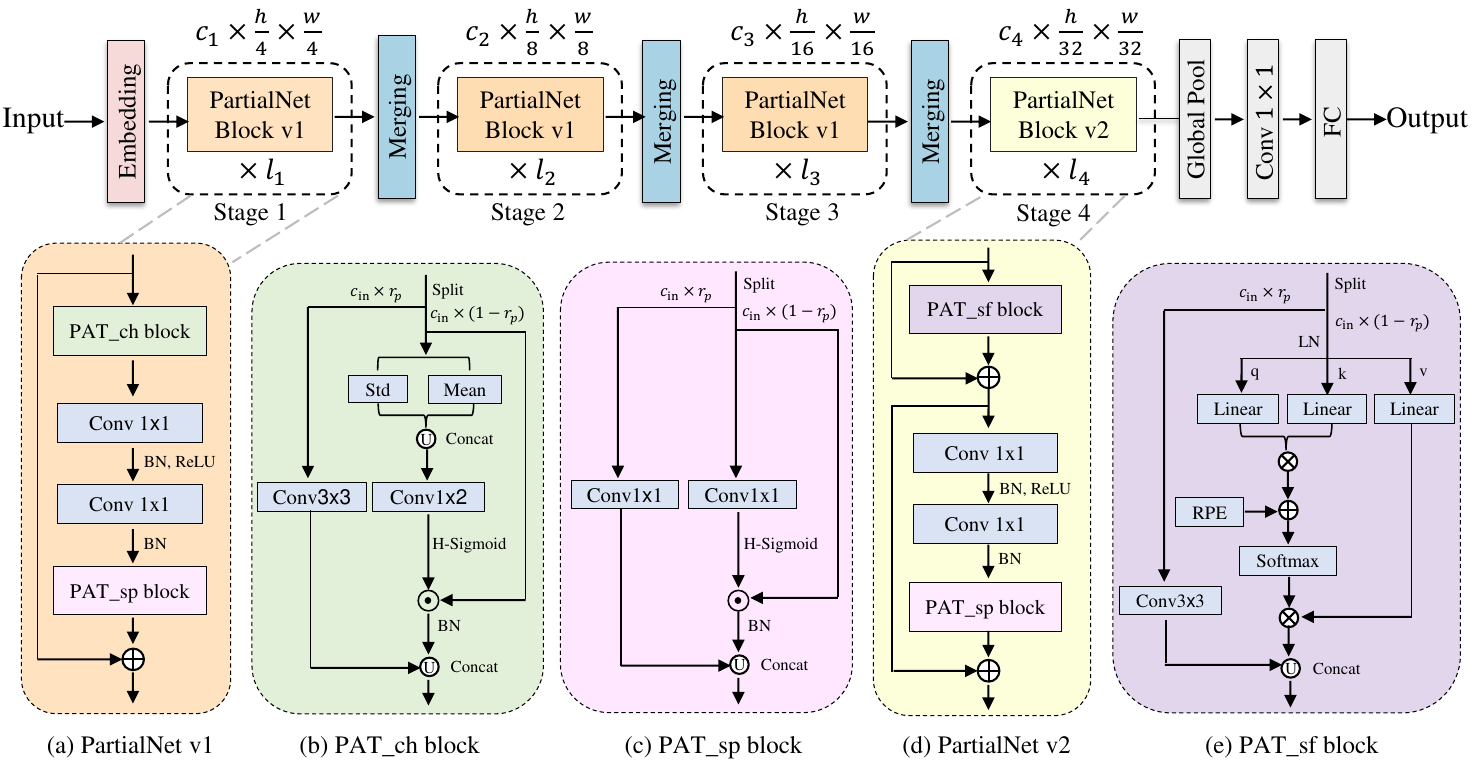}
  \caption{The overall architecture of our PartialNet, consisting of four hierarchical stages, each incorporating a series of PartialNet blocks followed by an embedding or merging layer. The last three layers are dedicated to feature classification. Where $\odot$ and $\otimes$ denote element-wise multiplication and matrix multiplication respectively.}
  \label{fig:network_overview}
\end{figure*}

In this section, we start by introducing the motivation behind enhancing partial convolution with visual attention and present our novel Partial Attention Convolution (PATConv) mechanism, which leverages attention on a subset of feature channels to balance computational efficiency and accuracy. Next, we detail three innovative blocks within PATConv: the Partial Channel-Attention block (PAT\_ch), which integrates Conv3$\times$3 with channel attention for global spatial interaction; the Partial Spatial-Attention block (PAT\_sp), which combines Conv1$\times$1 with spatial attention to efficiently mix channel-wise information; and the Partial Self-Attention block (PAT\_sf), which applies self-attention selectively to extend the model's receptive field. We further introduce a learnable dynamic partial convolution (DPConv) with adaptive channel split ratio for improved model flexibility. Finally, we describe the overall PartialNet architecture, structured in four hierarchical stages with PATConv-integrated blocks, aimed at achieving a robust speed-accuracy trade-off across model variants.

\subsection{Partial Channel Mechanism}
Generally, designing an efficient neural network necessitates comprehensive consideration and optimization from various perspectives, including fewer FLOPs, smaller model sizes, lower memory access, and better accuracy. Recently, some works (e.g., MobileViTv2 \cite{Mehta2022} and EfficientVit \cite{Cai2023b}) attempt to combine depthwise separable convolutions with the self-attention mechanism to reduce the model parameters and latency. Other works (e.g., ShuffleNetv2 \cite{Ma2018} and FasterNet \cite{Chen2023}) try to reduce FLOPs and improve inference speed by performing feature extraction using only a subset of the feature map channels. However, it does not exhibit a noticeable accuracy advantage when compared to models with similar parameters or FLOPs. Among, FasterNet only uses partial convolution, achieving exceptional speed across various devices. However, we find that FasterNet simply performs convolution operations on partial channels, which can reduce FLOPs and latency but leads to limited feature interaction and lack of global information exchange. 

In contrast, we comprehensively exploit the potential value within the channels of the entire feature map. For different partials, we use different operations to further reduce the model FLOPs while improving accuracy. It can be called the partial channel mechanism. Based on this, we propose a new type of convolution that replaces computationally expensive dense convolution operations with cost-effective visual attention, called Partial Attention Convolution (PATConv). Previous research~\cite{Han2020, Chen2023} has demonstrated that redundancy exists among feature channels, making attention operations applied to partial channels a form of global information interaction.

Unlike regular visual attention methods, our PATConv is more efficient due to using only a subset of channels for the computationally expensive element-wise multiplication. Indeed, running two operations in parallel on separate branches allows for simultaneous computation, optimizing resource utilization on the GPUs~\cite{kirk2016programming}. Suppose the input and output of our PATConv is denoted as {\small $F\in\mathbb{R}^{h\times w\times c_{in}}$} and {\small $O \in\mathbb{R}^{h\times w\times c_{out}}$} respectively, where {\small $c_{in}$} and {\small $c_{out}$} represent the number of input and output channels, {\small $h$}, {\small $w$} is the height and width of a channel, respectively. Suppose {\small $c_{in}=c_{out}$}, PATConv can be defined as
{\small
\begin{equation}
  O =\text{PATConv}(F) =\text{Conv}(F^{c_{in}\times r_p}) \cup \text{Atten}(F^{c_{in}(1-r_p)})
\end{equation}
}
where the symbol {\small $\cup$} and {\small Atten} denote the concatenation operation and the visual attention operation respectively. The {\small $r_p$} is a hyperparameter representing the split ratio of the channels and can be learned adaptively.

In addition, PATConv can apply channel-wise and spatial-wise mixing to enhance global information and integrate self-attention mechanisms to expand the model's receptive field to derive three blocks, proving to be highly effective.

\textbf{PAT\_ch:} We first propose to integrate Conv3$\times$3 and channel attention involving global spatial information interaction, and using an enhanced Gaussian-SE module compute channels' mean and variance to squeeze global spatial information. Unlike SENet~\cite{Hu2018}, it only considers the mean information of the channel and ignores the statistical information of std. Considering that the feature maps obey an approximately normal distribution~\cite{ioffe2015batch} during training, we fully utilize the Gaussian statistical to express the channel-wise representation information, as shown in~\cref{fig:network_overview}~(b).

\textbf{PAT\_sp:} Secondly, we integrate spatial attention with Conv1$\times$1 because both operations mix channel wise information. Our spatial attention employs a point-wise convolution to squeeze global channel information into tensor with only one channel. After passing through a Hard-Sigmoid activation, this tensor serves as the spatial attention map to weight features. We position PAT\_sp after the MLP layer, enabling the Conv1$\times$1 component of PAT\_sp to merge with the second Conv1$\times$1 in the MLP layer during inference, as shown in~\cref{fig:network_overview}~(c). This setup further minimizes the impact of attention on inference speed, and its details of the merge operation refer to Fig. 1~(d) in the appendix.

\textbf{PAT\_sf:} Finally, self-attention not only engages with spatial information interaction but also extends the model's effective receptive field, which can replace channel attention. However, because the computational complexity of self-attention operations increases quadratically with the size of the feature map, we restrict the use of PAT\_sf to the last stage to achieve a superior speed-accuracy trade-off. Beside, we employee relative position encoding (RPE)~\cite{Wu2021} into the attention map, which can further enhances model accuracy, as shown in~\cref{fig:network_overview}~(e).

Notably, unlike conventional CNNs combined with attention, which process steps one after the other, we process steps simultaneously on the same input, improving the balance between speed and accuracy. Moreover, our PATConv is not limited to the above three combinations, it can be efficiently combined with more visual attention modules. 

\subsection{Learnable Dynamic Partial Convolution}
For the PATConv, the split ratio {\small $r_p$} is a critical hyperparameter that significantly influences the parameters and latency of a model. A too-large {\small $r_p$} causes PATConv to degenerate into a regular convolution, rendering the visual attention component ineffective at capturing global information. Conversely, a too-small {\small $r_p$} results in PATConv lacking essential local inductive bias information.

\begin{figure}[ht]
  \centering
  \includegraphics[width=0.99\linewidth]{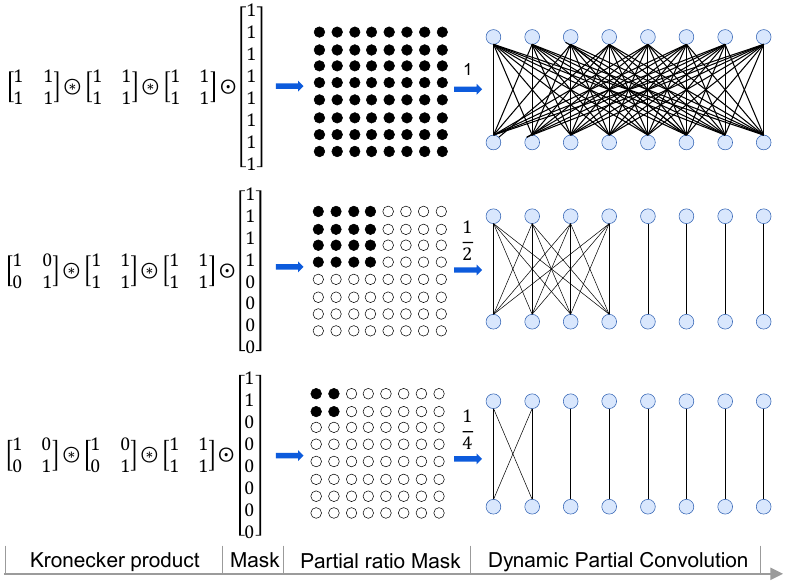}
  \caption{The generation process of DPConv, where $\odot$ denotes elementwise product, $\circledast $ denotes a Kronecker product.}
  \label{fig:DPConv_modul}
\end{figure}

Achieving higher accuracy and throughput at similar complexity often necessitates extensive experimentation to identify an optimal {\small $r_p$}. In FasterNet, a default split ratio of 1/4 is for all variants. In contrast, we propose a dynamic partial convolution (DPConv) in which the {\small $r_p$} is learnable. This approach allows a model to adaptively determine the optimal {\small $r_p$} for different layers during training. The strategies can be modeled by a binary relationship matrix {\small $U \in \{0,1\}^{c_{in}\times c_{out}}$}~\cite{Zhang2019c}. The entire matrix {\small $U$} can be decomposed in into a set of {\small $K$} small matrix {\small $U_{k} \in \{0,1\}^{2\times 2}$}, where {\small $U_{k}$} either equal to a 2-by-2 constant matrix of ones {\small $\mathbf{1}$} or equal to 2-by-2 identity matrix {\small $I$}, \ie, 
{\small
\begin{gather} 
 U =U_{1}\circledast U_{2}\circledast ... \circledast U_{K}\\
 U_{k} = g_{k}\mathbf{1}+(1-g_{k})I,\; \forall g_{k} \in g,\; g = \text{Sign}(\widetilde{g})
\end{gather}
}
where $\circledast$ denotes a Kronecker product~\cite{broxson2006kronecker} and {\small $\widetilde{g} \in \mathbb{R}^{K}$} is a learnable gate vector taking continues value, and {\small $g \in \{0,1\}^K$} is a binary gate vector derived from {\small $\widetilde{g}$}. Since the {\small Sign} function is not differentiable, the gate parameters are optimized using a straight-through estimator~\cite{Courbariaux2016}, similar to the binary network quantization method, to ensure convergence. Suppose {\small $c_{in}=c_{out}$}, so {\small $K=\log_{2}c_{in}$}. DPConv can be defined as
{\small
\begin{equation} 
  O =\text{DPConv}(F) = U \odot \mathbf{m} \odot W
\end{equation}
}
where $\odot$ denotes elementwise product, the {\small $\mathbf{m}\in \mathbb{R}^{c_{in}}$} is a vector to mask the useless part of {\small $U$}. Since the result of the Kronecker product is a power of 2 matrix, the number of channels in each layer of the model also needs to be {\small$2^K$}. So, it is not hard to deduce that 
{\small
\begin{equation} 
  r_p = \frac{2^{\sum_{k=1}^{K}(1-g_{k})}}{c_{in}},\; \mathbf{m}_i= 
  \begin{cases} 
  1 & \text{if } i < \frac{1}{r_p} \\
  0 & \text{if } i \geq \frac{1}{r_p}
  \end{cases} 
\end{equation}
}
where the {\small $g_{k}$} indicates the {\small $k$}-th component of {\small $g$}. The specific generation process of DPConv is shown in~\cref{fig:DPConv_modul}.

Considering the constraints of model deployment, \eg, the parameters and FLOPs, it is necessary to limit the learned {\small $r_p$} to avoid being too large. Therefore, we design a resource-constrained training scheme for DPConv. To simplify the calculation, assuming that we only consider the case of partial convolution. We propose a regularization term denoted as {\small $\zeta$} to constrain the computational complexity by
{\small
\begin{equation} 
  \zeta = \sum_{l=1}^{L}\zeta_l,\; \zeta_l=\sum_{i=1}^{c_{in}}\sum_{j=1}^{c_{out}}u_{ij}, \forall u_{ij} \in U 
  \label{eqn:DPConv_regu}
\end{equation}
}
where {\small $L$} denotes the number of DPConv layers, and {\small $u_{ij}$} denotes an element of {\small $U$}. The term {\small $\zeta_l$} represents the number of non-zero elements in {\small $U$}, measuring the number of activated convolution weights of the {\small $l$}-th DPConv layer. Thus, {\small $\zeta$} can be treated as a measurement of the model's computational complexity. In fact, it can be deduced that the sum of each row or each column of {\small $U$} can be calculated as {\small $\prod_{k=1}^{K}(1+g_k)$}. Substituting it to~\cref{eqn:DPConv_regu} gives us
{\small
\begin{equation}
  \zeta = \sum_{l=1}^{L} (\prod_{k=1}^{K^l}(1+g_k^l) \cdot \prod_{k=1}^{K^l}(1+g_k^l) ) = \sum_{l=1}^{L} 2^{2\cdot \sum_{k=1}^{K^l}g_k^l}
\end{equation}
}
where {\small $g_k^l$} and {\small $K^l$} indices {\small $g_k$} and {\small $K$} in the {\small $l$}-th layer, respectively. Here we suppose {\small $c_l=c_{in}=c_{out}$}. Let {\small $\kappa=\sum_{l=1}^{L}{\frac{c_l}{\theta}}^2$} represent the desire computational complexity of the entire network. By setting $\theta$, we can control the overall complexity of the PartialNet. For example, when {\small $\theta$}=4, which is equal to the complexity of {\small $r_p$}=1/4 in FasterNet. So, a weighted product {\small $[\frac{\kappa}{\zeta}]^\alpha$} to approximate the Pareto optimal problem, {\small $\alpha$} is a constant value by empirically set. And we have {\small $\alpha=0$} if {\small $\zeta \le \kappa$}, implying that the complexity constraint is satisfied. Otherwise, {\small $\alpha=-0.01$} is used to penalize the model complexity when {\small $\zeta > \kappa$}.

Additionally, the split ratio mask is closely related to the gate vector {\small $g$}. A reasonable Kronecker product can only be generated when the elements in {\small $g$} are ordered such that all elements of {\small $I$} come before those of {\small $\mathbf{1}$} for DPConv. Therefore, it is necessary to constrain {\small $g$} by incorporating a regularization loss term {\small $\psi$}, which can be computed by
{\small
\begin{equation} 
  \psi = \sum_{l=1}^{L}\psi_l,\; \psi_l=
  \begin{cases}
      \sum_{i}^{K}\left|\widetilde{g}_k^l\right| & \text{if } g^l_i=1\; \text{and}\; g^l_{i+1}=0, \\
      0 & \text{otherwise}.
  \end{cases}
\end{equation} 
}

Finally, our objective is to search a PartialNet model that
{\small
\begin{equation}
  \begin{cases}
    \text{minimize} \; &\mathcal{L}(\{w_l\}_{l=1}^L,\{\widetilde{g}_l\}_{l=1}^L)\cdot [\frac{\kappa}{\zeta}]^\alpha + \psi \beta , \\
    \text{subject to} \; &\zeta \le \kappa.
  \end{cases}
\end{equation} 
}
where {\small $\beta\in (0,1] $} is the penalty factor of {\small $\psi$}, default is 0.9.

\subsection{PartialNet Architecture}
The overall architecture of PartialNet is depicted in~\cref{fig:network_overview}, consists of four hierarchical stages, each of which precedes an embedding layer (a regular Conv4$\times$4 with stride 4) or a merging layer (a regular Conv2$\times$2 with stride 2). These layers serve for spatial downsampling and channel number expansion. Each stage comprises a set of PartialNet blocks. In the first three stages of the PartialNet, we employ ``PartialNet Block v1" including PAT\_ch block and PAT\_sp block, as shown in~\cref{fig:network_overview} (a). Similarly, we employ ``PartialNet Block v2" by replacing PAT\_ch with PAT\_sf in the last stage and modifying the shortcut connection way to achieve stable training, as shown in~\cref{fig:network_overview} (d). 

In addition, we maintain normalization or activation layers only after each intermediate Conv1$\times$1 to preserve feature diversity and achieve higher throughput. We also incorporate batch normalization into adjacent Conv layers to expedite inference without sacrificing performance. For the activation layer, the smaller PartialNet variants uses GELU, while the larger PartialNet variants employs ReLU. The last three layers consist of global average pooling, Conv1$\times$1, and a fully connected layer. These layers collectively serve for feature transformation and classification. We offer tiny, small, medium, and large variants of PartialNet, which are denoted as PartialNet-T0/1/2, PartialNet-S, PartialNet-M, and PartialNet-L. These variants share a similar architecture but differ in depth and width. For detailed specifications please refer to Tab. 3 of the appendix. 

\begin{table*}[ht] \small
  \centering
  \resizebox{1.0\linewidth}{!}{ 
  \begin{tabular}{@{}lcccccccc@{}}
    \toprule
    Network   &\makecell{Type}   & \makecell{Params (M)$\downarrow$}	&\makecell{FLOPs (G)$\downarrow$}	  &\makecell{TP V100 (FPS)$\uparrow$} &\makecell{TP MI250 (FPS)$\uparrow$}	&\makecell{Latency CPU (ms)$\downarrow$} &\makecell{Top-1 (\%)$\uparrow$}  \\
    \hline
    ShuffleNetV2 x1.5\cite{Ma2018}      &CNN      & 3.5       & 0.30  &5315       & 6642        & 13.7        & 72.6       \\
    MobileNetV2\cite{Sandler2018a}      &CNN      & 3.5       & 0.31  &3924       & 7359        & 13.7        & 72.0       \\
    FasterNet-T0\cite{Chen2023}         &CNN      & 3.9       & 0.34  &{\bf 8546} & 10612       & {\bf 10.5}  & 71.9       \\
    MobileViTv2-0.5\cite{Mehta2022}     &Hybrid   & 1.4       & 0.46  &3094       & 3135        & 15.8        & 70.2       \\
    PartialNet-T0(ours)                 &Hybrid   & 4.3       & 0.25  &7777       & {\bf 11744} & 12.2        & {\bf 73.9} \\
    \hline
    EfficientNet-B0\cite{Tan2019}       &CNN      & 5.3       & 0.39  &2934       & 3344        & 22.7        & 77.1       \\
    ShuffleNetV2 x2\cite{Ma2018}        &CNN      & 7.4       & 0.59  &4290       & 5371        & 22.6        & 74.9       \\
    MobileNetV2 x1.4\cite{Sandler2018a} &CNN      & 6.1       & 0.60  &2615       & 4142        & 21.7        & 74.7       \\
    FasterNet-T1\cite{Chen2023}         &CNN      & 7.6       & 0.85  &{\bf 4648} & 7198        & 22.2        & 76.2       \\
    PartialNet-T1(ours)                 &Hybrid   & 7.8       & 0.55  &4403       & {\bf 7379}  & {\bf 21.5}  & {\bf 78.1} \\
    \hline
    EfficientNet-B1\cite{Tan2019}       &CNN      & 7.8       & 0.70  &1730       & 1583        & 35.5        & 79.1       \\
    ResNet50\cite{He2015}               &CNN      & 25.6      & 4.11  &1258       & 3135        & 94.8        & 78.8       \\
    FasterNet-T2\cite{Chen2023}         &CNN      & 15.0      & 1.91  &2455       & 4189        & 43.7        & 78.9       \\
    PoolFormer-S12\cite{Yu2022a}        &Hybrid   & 11.9      & 1.82  &1927       & 3558        & 56.1        & 77.2       \\
    MobileViTv2-1.0\cite{Mehta2022}     &Hybrid   & 4.9       & 1.85  &1391       & 1543        & 41.5        & 78.1       \\
    EfficientViT-B1\cite{Cai2023b}      &Hybrid   & 9.1       & 0.52  &3072       & 3387        & {\bf 25.7}  & 79.4       \\
    PartialNet-T2(ours)                 &Hybrid   & 12.6      & 1.03  &{\bf 3074} & {\bf 4761}  & 35.2        & {\bf 80.2} \\
    \hline
    EfficientNet-B3\cite{Tan2019}       &CNN      & 12.0      & 1.80  &768        & 926         & 73.5        & 81.6       \\
    FasterNet-S\cite{Chen2023}          &CNN      & 31.1      & 4.56  &1261       & 2243        & 96.0        & 81.3       \\
    PoolFormer-S36\cite{Yu2022a}        &Hybrid   & 30.9      & 5.00  &675        & 1092        & 152.4       & 81.4       \\
    MobileViTv2-2.0\cite{Mehta2022}     &Hybrid   & 18.5      & 7.50  &551        & 684         & 103.7       & 81.2       \\
    Swin-T\cite{Liu2021c}               &Hybrid   & 28.3      & 4.51  &808        & 1192        & 107.1       & 81.3       \\
    PartialNet-S(ours)                  &Hybrid   & 29.0      & 2.71  &{\bf 1559} & {\bf 2422}  & {\bf 72.5}  & {\bf 82.1} \\
    \hline
    EfficientNet-B4\cite{Tan2019}       &CNN      & 19.0      & 4.20  &356        & 442         & 156.9       & 82.9       \\
    FasterNet-M\cite{Chen2023}          &CNN      & 53.5      & 8.74  &621        & 1098        & 181.6       & 83.0       \\
    PoolFormer-M36\cite{Yu2022a}        &Hybrid   & 56.2      & 8.80  &444        & 721         & 244.3       & 82.1       \\
    Swin-S\cite{Liu2021c}               &Hybrid   & 49.6      & 8.77  &477        & 732         & 199.1       & 83.0       \\
    PartialNet-M(ours)                  &Hybrid   & 61.3      & 6.69  &{\bf 799}  & {\bf 1280}  & {\bf 155.3} & {\bf 83.1} \\
    \hline
    EfficientNet-B5\cite{Tan2019}       &CNN      & 30.0      & 9.90  &246        & 313         & 333.3       & 83.6       \\
    ConvNeXt-B\cite{Liu2022b}           &CNN      & 88.6      & 15.38 &322        & 430         & 317.1       & 83.8       \\
    FasterNet-L\cite{Chen2023}          &CNN      & 93.5      & 15.52 &384        & 709         & 312.5       & 83.5       \\
    PoolFormer-M48\cite{Yu2022a}        &Hybrid   & 73.5      & 11.59 &335        & 556         & 322.3       & 82.5       \\
    Swin-B\cite{Liu2021c}               &Hybrid   & 87.8      & 15.47 &315        & 520         & 333.8       & 83.5       \\
    PartialNet-L(ours)                  &Hybrid   & 104.3     & 11.91 &{\bf 426}  & {\bf 765}   & {\bf 272.5} & {\bf 83.9} \\
    \bottomrule
  \end{tabular}
  }
  \caption{Comparison on ImageNet-1K Benchmark: Models with similar top-1 accuracy are grouped together. The TP denotes throughput.}
  \label{tab:class_Benchmark}
\end{table*}

\section{Experiments}
\label{sec:experiments}
\subsection{PartialNet on ImageNet-1k Classification}
{\bf Setup.} ImageNet-1K is one of the most extensively used datasets in computer vision. It encompasses 1K common classes, consisting of approximately 1.3M training images and 50K validation images. We train our models on the ImageNet-1k dataset for 300 epochs using AdamW optimizer with 20 epochs linear warm-up. And we use the same regularization and augmentation techniques and multi-scale training as FasterNet. For detailed experimental settings please refer to Tab. 2 of the appendix. For inference speed, we test the model's throughput in Nvidia-V100 and AMD-Instinct-MI250 GPUs with batch size of 256, we test latency in AMD $\texttt{EPYC}^{TM}$ 73F3 CPU with one core.  

{\bf Results.} \cref{tab:class_Benchmark} provides a comparison of our PartialNet models (T0, T1, T2, S, M, and L) with previous SOTA cnn-based and hybrid-based models. The experimental results demonstrate that PartialNet consistently surpasses recent models like FasterNet~\cite{Chen2023} across all model variants. For example, PartialNet-T2 achieves 1.3\% higher accuracy than FasterNet-T2 while exhibiting around 25.2\%(or 13.7\%) increase in V100(or MI250) throughput and 24.1\% lower CPU latency. The results demonstrate that the combination of partial visual attention and partial convolution significantly improves model performance while increasing throughput. There is a certain difference in hardware architecture between the V100 and MI250 because the V100 is better suited for compute-intensive tasks, while the MI250 excels in bandwidth-intensive tasks~\cite{Halbiniak2024}. This may explain why our PartialNet has slightly lower throughput than FasterNet in the T0 and T1 variants on the V100. For more comparison please refer to Tab. 4 of the appendix. 

\begin{figure}[ht]
  \centering
  \includegraphics[width=0.95\linewidth]{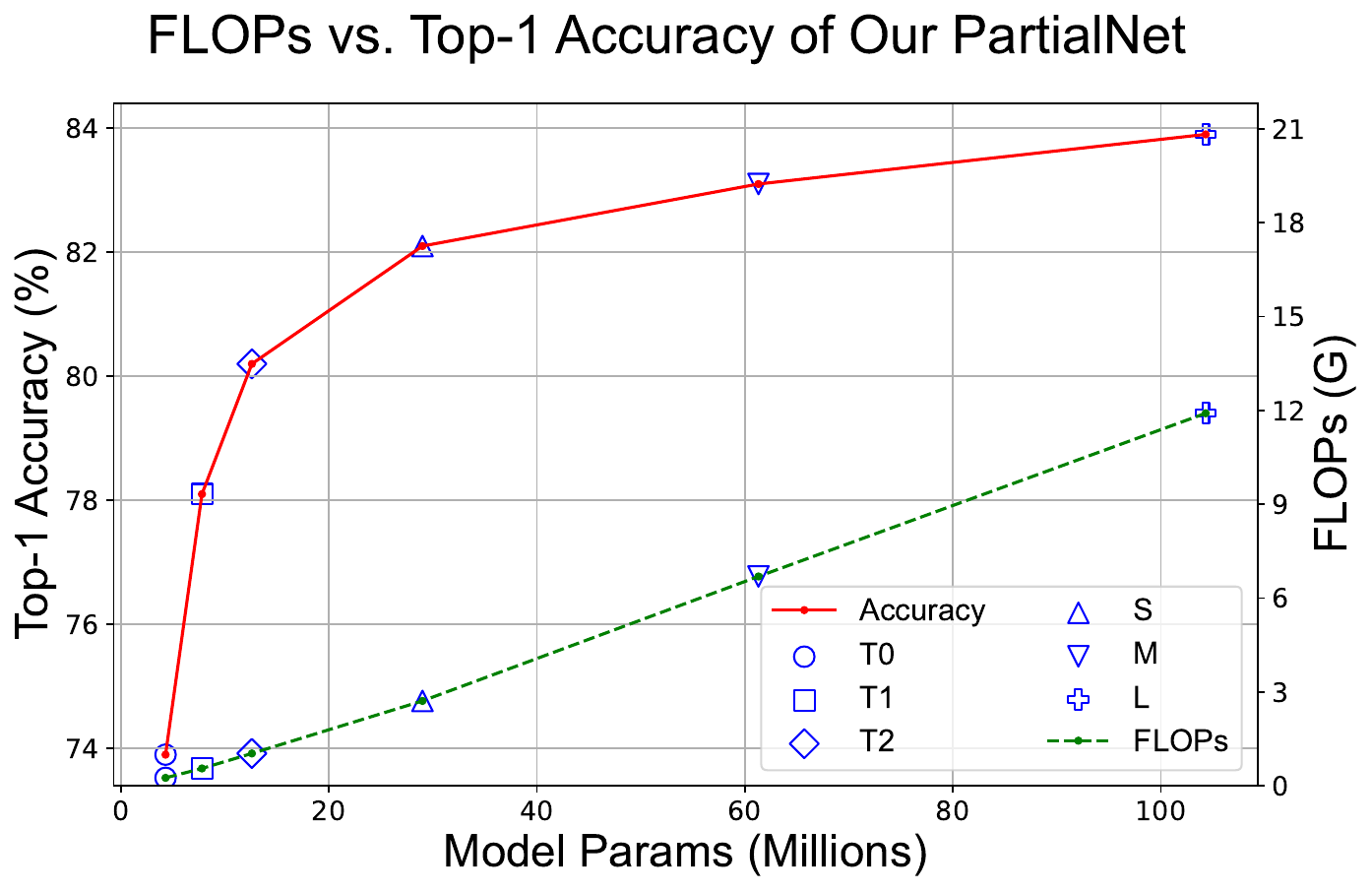}
  \caption{The relationship between FLOPs and Top-1 Accuracy in different PartialNet model variants.}
  \label{fig:FLOPs_vs_Accuracy}
\end{figure}

In addition, it can be seen from~\cref{fig:FLOPs_vs_Accuracy}, as our PartialNet variants gradually increase from T0 to L, both Top-1 accuracy and FLOPs increase, but the improvement in Top-1 accuracy relative to the increase in FLOPs is more pronounced. This further demonstrates that our PartialNet achieves better accuracy with less computational cost.

\subsection{PartialNet on Downstream Tasks}

\begin{table*}[ht] \small
  \centering
  \resizebox{1.0\linewidth}{!}{
  \begin{tabular}{@{}lccccc cccc@{}}
    \toprule
    Backbone              & \makecell{Params (M)$\downarrow$}	&\makecell{FLOPs (G)$\downarrow$} &\makecell{Throughput (FPS)$\uparrow$}	&\makecell{$AP^{b}\uparrow$}  &\makecell{$AP^{b}_{50}\uparrow$} &\makecell{$AP^{b}_{75}\uparrow$} &\makecell{$AP^{m}\uparrow$} &\makecell{$AP^{m}_{50}\uparrow$} &\makecell{$AP^{m}_{75}\uparrow$}\\
    \hline
    ResNet50\cite{He2015}                 & 44.2    & 253   &121         & 38.0       & 58.6       & 41.4       & 34.4       & 55.1       & 36.7       \\
    PoolFormer-S24\cite{Yu2022a}          & 41.0    & 233   &68          & 40.1       & 62.2       & 43.4       & 37.0       & 59.1       & 39.6       \\
    PVT-Small x1.5\cite{Wang2021c}        & 44.1    & 238   &98          & 40.4       & 62.9       & 43.8       & 37.8       & 60.1       & 40.3       \\
    FasterNet-S\cite{Chen2023}            & 49.0    & 258   &121         & 39.9       & 61.2       & 43.6       & 36.9       & 58.1       & 39.7       \\
    PartialNet-S(ours)                    & 46.9    & 216   &{\bf 122}   & {\bf 42.7} & {\bf 64.9} & {\bf 46.5} & {\bf 39.3} & {\bf 61.8} & {\bf 42.2} \\
    \hline
    ResNet101\cite{Mehta2021}             & 63.2    & 329   &62          & 40.4       & 61.1       & 44.2       & 36.4       & 57.7       & 38.8       \\
    ResNeXt101-32$\times$4d\cite{Xie2017} & 62.8    & 333   &51          & 41.9       & 62.5       & 45.9       & 37.5       & 59.4       & 40.2       \\
    PoolFormer-S36\cite{Yu2022a}          & 50.5    & 266   &44          & 41.0       & 63.1       & 44.8       & 37.7       & 60.1       & 40.0       \\
    PVT-Medium\cite{Wang2021c}            & 63.9    & 295   &52          & 42.0       & 64.4       & 45.6       & 39.0       & 61.6       & 42.1       \\
    FasterNet-M\cite{Chen2023}            & 71.2    & 344   &62          & 43.0       & 64.4       & 47.4       & 39.1       & 61.5       & 42.3       \\
    PartialNet-M(ours)                    & 78.2    & 295   &{\bf 65}    & {\bf 44.3} & {\bf 65.8} & {\bf 48.5} & {\bf 40.6} & {\bf 63.3} & {\bf 43.7} \\
    \hline
    ResNeXt101-64$\times$4d\cite{Xie2017} & 101.9   & 487   &29          & 42.8       & 63.8       & 47.3       & 38.4       & 60.6       & 41.3       \\
    PVT-Large$\times$4d\cite{Wang2021c}   & 81.0    & 358   &26          & 42.9       & 65.0       & 46.6       & 39.5       & 61.9       & 42.5       \\
    FasterNet-L\cite{Chen2023}            & 110.9   & 484   &35          & 44.0       & 65.6       & 48.2       & 39.9       & 62.3       & 43.0       \\
    PartialNet-L(ours)                    & 122.0   & 397   &{\bf 39}    & {\bf 44.7} & {\bf 66.3} & {\bf 49.0} & {\bf 41.0} & {\bf 63.7} & {\bf 44.2} \\
    \bottomrule
  \end{tabular}
  }
  \caption{Results using PartialNet-S/M/L on object detection and instance segmentation benchmark in COCO dataset.}
  \label{tab:dect_and_seg_Benchmark}
\end{table*}

{\bf Setup.} We utilize the pre-trained PartialNet as the backbone within the Mask-RCNN~\cite{he2017mask} detector for object detection and instance segmentation on the MS-COCO 2017 dataset, comprising 118K training images and 5K validation images. To highlight the effectiveness of the backbone itself, we follow the FasterNet approach and employ the AdamW optimizer, conduct training of 12 epochs, use a batch size of 16, image size of 1333$\times$800, and maintain other training settings without further hyperparameter tuning.

{\bf Results.} \cref{tab:dect_and_seg_Benchmark} presents a comparison of PartialNet with representative models, reporting performance in terms of average precision (mAP) for both detection and instance segmentation. The results show that PartialNet consistently outperforms mainstream SOTA models, achieving higher Average Precision (AP) while maintaining similar inference speed. For instance, PartialNet-S achieves over 7\% higher $AP^{b}$ in detection metrics and 6.5\% higher $AP^{m}$ in segmentation metrics with fewer parameters and FLOPs compared to FasterNet-S, while maintaining similar throughput. The results further confirm the generalization capabilities of our proposed PartialNet across various tasks.

\subsection{Ablation Studies}
\label{sec:ablation}
\begin{figure}[ht]
  \centering
  \includegraphics[width=0.98\linewidth]{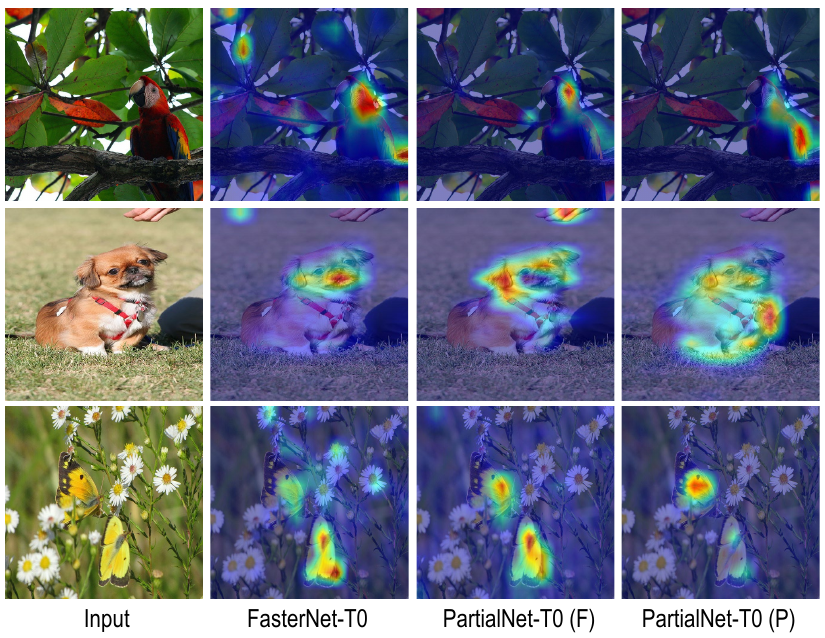}
  \caption{Visualization results show different categories of the ImageNet-1K validation set using Grad-CAM.}
  \label{fig:cam_atten}
\end{figure}

~~~\textbf{Partial Attention vs. Full Attention.}
To prove the superiority of our PATConv over full attention, we conduct comparative experiments on the PartialNet-T2, as shown in~\cref{tab:abla_block_acc}. Specifically, we replace PATConv with corresponding regular full attention for comparison. Full attention involves conducting visual attention calculations on all channels of the input feature map, which is the common way of conventional visual attention mechanism. The results indicate that our PATConv achieves a superior balance between inference speed and performance compared to the full attention counterpart. In addition, we adopt Grad-CAM~\cite{Selvaraju2017} to visualize the attention. Results in~\cref{fig:cam_atten} show that partial visual attention can focus on the target objects. It is feasible to perform attention operations on part channels and confirm the effectiveness of our improved partial channel attention mechanism.

\begin{table}[ht] \small
  \centering
  \resizebox{1.0\linewidth}{!}{
  \begin{tabular}{@{}lcc ccc@{}}
    \toprule
    \makecell{ch-sp-sf} & \makecell{Params\\(M)} & \makecell{FLOPs\\(G)} & \makecell{Throughput\\(FPS)$\uparrow$} & \makecell{Latency\\(ms)$\downarrow$} & \makecell{Top-1\\(\%)$\uparrow$} \\
    \hline
    P-P-P     & 12.6       & 1.03     & {\bf 4761}   & {\bf 35.2}  & {\bf 80.2}                          \\
    F-P-P     & 13.0       & 1.04     & 4662         & 36.5        & 80.1                          \\
    P-F-P     & 12.6       & 1.04     & 4688         & 35.6        & 79.9                          \\
    P-P-F     & 14.5       & 1.12     & 4600         & 38.6        & {\bf 80.2}                         \\
    \bottomrule
  \end{tabular}
  }
  \caption{Comparison on PartialNet-T2 of partial attention (P), and full attention (F) on ImageNet-1K dataset. Where the ``ch", ``sp", and ``sf" denote channel-wise attention, spatial-wise attention, and self-attention respectively.}
  \label{tab:abla_block_acc}
\end{table}

\textbf{Effect of three PATConv blocks.}
To confirm the individual effects of our proposed three PATConv blocks, we conducted ablation studies by progressively adding each block one by one, as indicated in~\cref{tab:abla_PAT_acc}. The results indicate that the three proposed PATConv blocks consistently enhance model performance. Additionally, \cref{tab:abla_block_acc_other} also provides a reproduced comparison of the PATConv applied to other models. The results further demonstrate the effectiveness of our proposed PATConv blocks. The training process of ConvNext-tiny is shown in Fig. 2 of the appendix.

\begin{table}[ht] \small
  \centering
  \resizebox{1.0\linewidth}{!}{
  \begin{tabular}{@{}lccc cccc@{}}
    \toprule
    \makecell{ch} & \makecell{sp} & \makecell{sf} & \makecell{Params\\(M)} & \makecell{FLOPs\\(G)} & \makecell{Throughput\\(FPS)$\uparrow$} & \makecell{Latency\\(ms)$\downarrow$} & \makecell{Top-1\\(\%)$\uparrow$} \\
    \hline
    w/o        & w/o          & w/o       &11.1  &0.92      & {\bf 6405} & {\bf 25.7} & 76.0                          \\
    w.         & w/o          & w/o       &11.1  &0.92      & 5440       & 30.9       & 77.4                          \\
    w.         & w.           & w/o       &11.5  &0.92      & 5157       & 31.7       & 78.9                          \\
    w.         & w.           & w.        &12.6  &1.03      & 4761       & 35.2       & {\bf 80.2}                    \\
    \bottomrule
  \end{tabular}
  }
  \caption{Ablation experiments of PartialNet-T2 with different configurations of PAT blocks on the ImageNet-1K.}
  \label{tab:abla_PAT_acc}
\end{table}

\begin{table}[ht] \small
  \centering
  \resizebox{1.0\linewidth}{!}{
  \begin{tabular}{@{}lcc |cc@{}}
    \toprule
    \multirow{2}{*}{Model}  & \multicolumn{2}{c}{Throughput(FPS)$\uparrow$} & \multicolumn{2}{c}{Top-1(\%)$\uparrow$}  \\
    \cmidrule(lr){2-5}      &  original    & PATConv    & original    &   PATConv     \\
    \hline
    ResNet50                &   1258       &  2832      &   76.13     &  77.64       \\
    MobileNetV2             &   3924       &  4560      &   71.14     &  73.85        \\
    ConvNext-tiny           &   902        &  1123      &   76.00     &  78.57        \\
    \bottomrule
  \end{tabular}
  }
  \caption{The results of applying PATConv to other models.}
  \label{tab:abla_block_acc_other}
\end{table}

\textbf{PATConv vs. Regular Conv and DWConv.}
To further verify the advantages of our proposed partial attention convolution, such as PAT\_ch, over regular convolution (Conv) and DepthWise convolution (DWConv), we conduct ablation experiments on PartialNet-T2, as is shown in~\cref{tab:abla_PAT_ch}. To make a fair comparison, we widen DWConv to keep the throughput of the three convolution types in the same range. The results show that our proposed PAT\_ch surpasses Conv and DWConv in all metrics including Params, Flops, throughput, latency and Top-1 accuracy, which validates the efficiency and effectiveness of PATConv. 

\begin{table}[ht] \small
  \centering
  \resizebox{1.0\linewidth}{!}{
  \begin{tabular}{@{}lcccc cc@{}}
    \toprule
    \makecell{Conv3$\times$3} & \makecell{Params\\(M)} & \makecell{FLOPs\\(G)} & \makecell{Throughput\\(FPS)$\uparrow$} & \makecell{Latency\\(ms)$\downarrow$} & \makecell{Top-1 (\%)$\uparrow$} \\
    \hline
    PAT\_ch        & {\bf 12.6}             & {\bf 1.03}           & {\bf 4761}         & {\bf 35.2}        & {\bf 80.2}                          \\
    Conv           & 15.8             & 2.12           & 4190               & 49.9               & 79.9                          \\
    DWConv         & 15.8             & 1.28           & 4017               & 35.4               & 79.6                          \\
    \bottomrule
  \end{tabular}
  }
  \caption{Ablation on PartialNet-T2 with different convolution types on ImageNet-1K dataset.}
  \label{tab:abla_PAT_ch}
\end{table}

\begin{figure}[ht]
  \centering
  \includegraphics[width=1.0\linewidth]{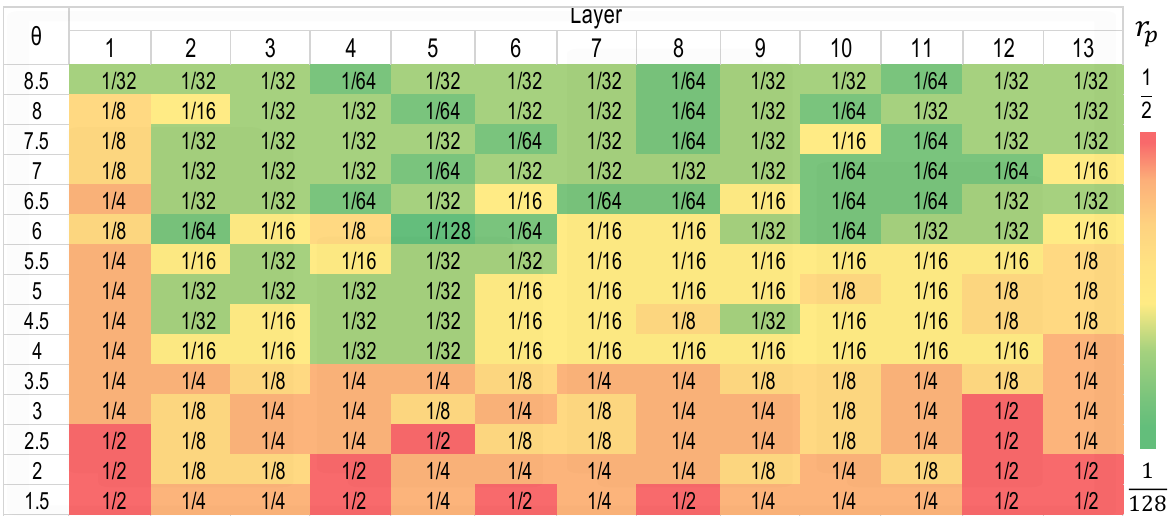}
  \caption{The learned split ratios {\small $r_p$} of different layers of PartialNet-T0 under different complexity constraints {\small $\theta$}.}
  \label{fig:abla_ratio}
\end{figure}

\textbf{Different Constraints Settings.}
For different constraints {\small $\theta$} and other parameters fixed, we conduct different experiments on PartialNet-T0, as is shown in~\cref{fig:abla_ratio}. We find that the learned split ratio {\small $r_p$} across different layers exhibit a consistent pattern: as the {\small $\theta$} increase, the first and last layers are less likely to exhibit sparsity compared to the middle layers, which is consistent with the conclusion of model quantization~\cite{Dong2019}, that the first and last layers of the model are more important. For detailed specifications of split ratio {\small $r_p$} across different layers please refer to Tab. 3 of the appendix.

\section{Conclusion}
\label{sec:conclusion}
Feature selection theory shows that there may be a certain degree of redundancy and correlation between features. While this redundancy does not provide additional information gain, it can increase computational complexity and heighten the risk of overfitting. Our research builds on this theory from an implementation perspective, achieving a balance of optimal performance and computational efficiency. Specifically, we introduce the partial channel mechanism and propose Partial Attention Convolution, which strategically integrates visual attention into the convolution process to enhance feature utility. Furthermore, we present Dynamic Partial Convolution, an adaptive approach that learns optimal split ratios for channels across different layers in the model. With these innovations, we develop the PartialNet architecture, which surpasses recent efficient networks on ImageNet-1K classification as well as COCO detection and segmentation tasks. This underscores the effectiveness of the partial channel mechanism in achieving an optimal balance between accuracy and efficiency across a range of vision tasks.
{
    \small
    \bibliographystyle{ieeenat_fullname}
    \bibliography{main}
}

\clearpage
\setcounter{page}{1}
\maketitlesupplementary

\label{sec:appandix}
\begin{figure*}[ht]
  \centering
  \includegraphics[width=0.83\textwidth]{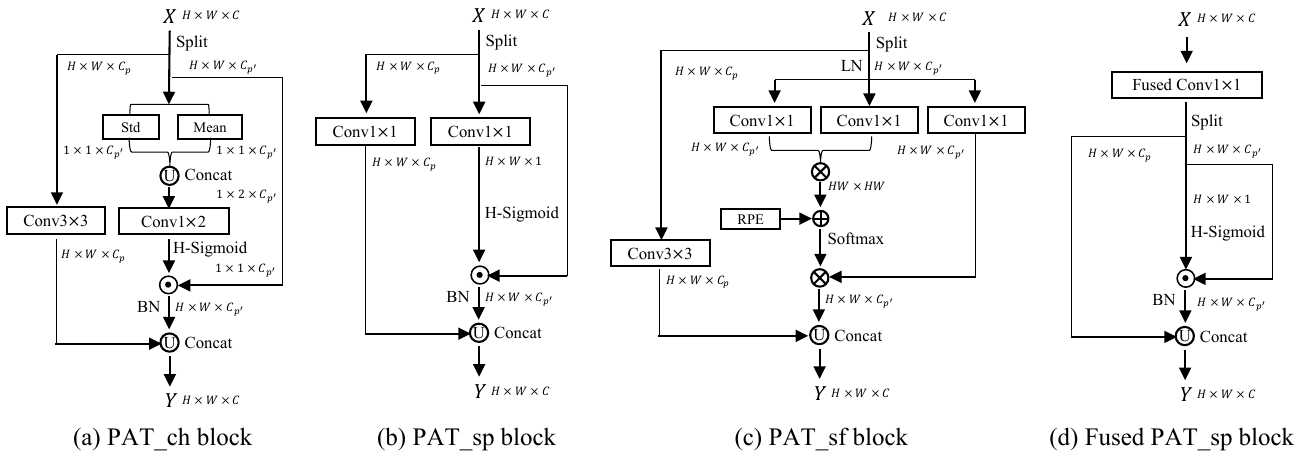}
  \caption{Detailed of three PATConv blocks. Where $\odot$ and $\otimes$ denote element-wise multiplication and matrix multiplication respectively, and $C=C_p+C_{p^{'}}$.}
  \label{fig:attention_block}
\end{figure*}

\begin{figure*}[ht]
  \centering
  \includegraphics[width=0.82\textwidth]{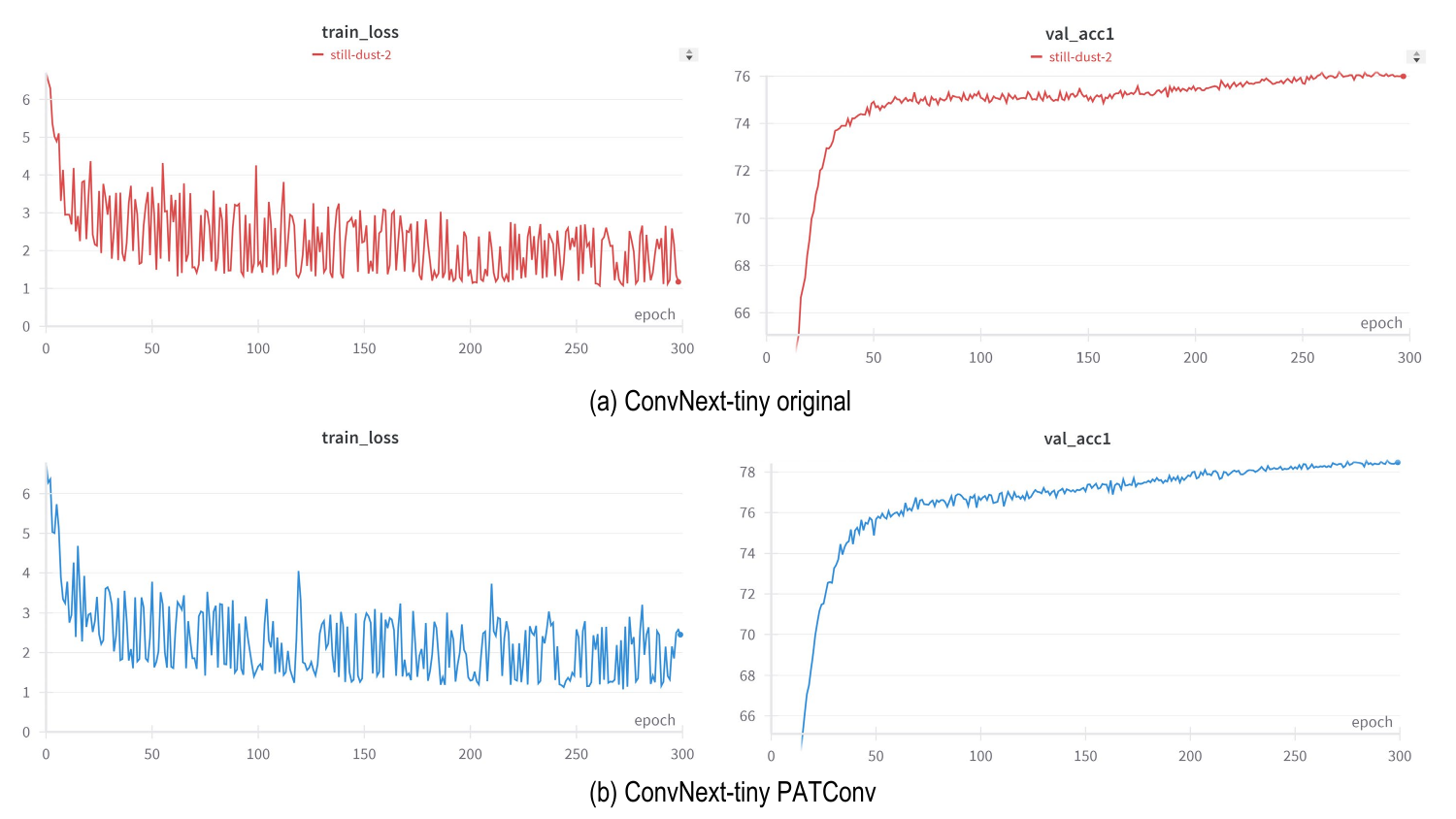}
  \caption{The training process of ConvNext-tiny with and without PATConv (i.e., PAT\_ch).}
  \label{fig:abla_convnext}
\end{figure*}

\section{Overview}
In this supplementary material, we present more explanations and experimental results.
\begin{itemize}
  \item Firstly, we provide detailed explanations of our experimental setup, the specifics of the three PATConv blocks, and the different PartialNet variants.
  \item Secondly, we present a comprehensive comparison of the classification task on the ImageNet-1k benchmark, as well as object detection and instance segmentation tasks on the COCO dataset.
  \item Finally, we provide additional ablation studies for our proposed Partial Attention Convolution (PATConv) and show the training process of ConvNext-tiny with and without PATConv (i.e., PAT\_ch).
\end{itemize}

\begin{table}[ht]
  \small
  \centering
  \resizebox{1\linewidth}{!}{%
    \setlength{\tabcolsep}{6pt}
    \begin{tabular}{@{}l|ccc@{}}
      \toprule
      Variants           & \qquad S \qquad \qquad                                            & \qquad M \qquad \qquad & L      \\
      \midrule
      Train and test Res & \multicolumn{3}{c}{shorter side $=$ 800, longer side $\leq$ 1333}                                   \\
      Batch size         & \multicolumn{3}{c}{16 (2 on each GPU)}                                                              \\
      Optimizer          & \multicolumn{3}{c}{AdamW}                                                                           \\
      Train schedule     & \multicolumn{3}{c}{1$\times$ schedule (12 epochs)}                                                  \\
      Weight decay       & \multicolumn{3}{c}{0.0001}                                                                          \\
      Warmup schedule    & \multicolumn{3}{c}{linear}                                                                          \\
      Warmup iterations  & \multicolumn{3}{c}{500}                                                                             \\
      LR decay           & \multicolumn{3}{c}{StepLR at epoch 8 and 11 with decay rate 0.1}                                    \\
      LR                 & 0.0002                                                            & 0.0001                 & 0.0001 \\
      Stoch. Depth       & 0.15                                                              & 0.2                    & 0.3    \\
      \bottomrule
    \end{tabular}%
  }
  \caption{Experimental settings of object detection and instance segmentation on the COCO2017 dataset.}
  \vspace{-0.2in}
  \label{tab:coco_settings}
\end{table}

\section{Clarifications on Experimental Setting}
Firstly, we provide the ImageNet-1k training and evaluation settings in~\cref{tab:imagenet_settings}. These settings can be used to reproduce our main results in Figure 1 of the main paper. Different PartialNet variants vary in the magnitude of regularization and augmentation techniques. The magnitude increases as the model size increases to alleviate overfitting and improve accuracy. It is worth noting that most of the works compared in Figure 1 of the main paper, such as MobileViT, FastNet, ConvNeXt, Swin, etc., also adopt such advanced training techniques (ADT), with some even heavily relying on hyper-parameter search. For other models without ADT, such as ShuffleNetV2, MobileNetV2, and GhostNet, although the comparison is not entirely fair, we include them for reference. Moreover, for object detection and instance segmentation on the COCO2017 dataset, we equip our PartialNet backbone with the popular Mask R-CNN detector. We use ImageNet-1k pre-trained weights to initialize the backbone and Xavier initialization for the add-on layers. Detailed settings are summarized in~\cref{tab:coco_settings}.

\begin{table}[ht]
  \small
  \centering
  \resizebox{1\linewidth}{!}{
    \begin{tabular}{@{}l|cccccc@{}}
      \toprule
      Variants           & T0                                & T1           & T2          & S           & M       & L     \\
      \hline
      Train Res          & \multicolumn{6}{c}{Random select from \{128,160,192,224,256,288\}}                             \\
      Test Res           & \multicolumn{6}{c}{224}                                                                        \\
      \hline
      Epochs             & \multicolumn{6}{c}{300}                                                                        \\
      \# of forward pass & \multicolumn{6}{c}{188k}                                                                       \\
      \hline
      Batch size         & 4096                              & 4096         & 4096        & 4096        & 2048    & 2048  \\
      Optimizer          & \multicolumn{6}{c}{AdamW}                                                                      \\
      Momentum           & \multicolumn{6}{c}{0.9/0.999}                                                                  \\
      LR                 & 0.004                             & 0.004        & 0.004       & 0.004       & 0.002   & 0.002 \\
      LR decay           & \multicolumn{6}{c}{cosine}                                                                     \\
      Weight decay       & 0.005                             & 0.01         & 0.02        & 0.03        & 0.05    & 0.05  \\
      Warmup epochs      & \multicolumn{6}{c}{20}                                                                         \\
      Warmup schedule    & \multicolumn{6}{c}{linear}                                                                     \\
      \hline
      Label smoothing    & \multicolumn{6}{c}{0.1}                                                                        \\
      Dropout            & \multicolumn{6}{c}{\ding{55}}                                                                  \\
      Stoch. Depth       & \ding{55}                         & 0.02         & 0.05        & 0.1         & 0.2     & 0.3   \\
      Repeated Aug       & \multicolumn{6}{c}{\ding{55}}                                                                  \\
      Gradient Clip.     & \ding{55}                         & \ding{55}    & \ding{55}   & \ding{55}   & 1       & 0.01  \\
      \hline
      H. flip            & \multicolumn{6}{c}{\ding{51}}                                                                  \\
      RRC                & \multicolumn{6}{c}{\ding{51}}                                                                  \\
      Rand Augment       & \ding{55}                         & 3/0.5        & 5/0.5       & 7/0.5       & 7/0.5   & 7/0.5 \\
      Auto Augment       & \multicolumn{6}{c}{\ding{55}}                                                                  \\
      Mixup alpha        & 0.05                              & 0.1          & 0.1         & 0.3         & 0.5     & 0.7   \\
      Cutmix alpha       & \multicolumn{6}{c}{1.0}                                                                        \\
      Erasing prob.      & \multicolumn{6}{c}{\ding{55}}                                                                  \\
      Color Jitter       & \multicolumn{6}{c}{\ding{55}}                                                                  \\
      PCA lighting       & \multicolumn{6}{c}{\ding{55}}                                                                  \\
      \hline
      SWA                & \multicolumn{6}{c}{\ding{55}}                                                                  \\
      EMA                & \multicolumn{6}{c}{\ding{55}}                                                                  \\
      \hline
      Layer scale        & \multicolumn{6}{c}{\ding{55}}                                                                  \\
      \hline
      CE loss            & \multicolumn{6}{c}{\ding{51}}                                                                  \\
      BCE loss           & \multicolumn{6}{c}{\ding{55}}                                                                  \\
      \hline
      Mixed precision    & \multicolumn{6}{c}{\ding{51}}                                                                  \\
      \hline
      Test crop ratio    & \multicolumn{6}{c}{0.9}                                                                        \\
      \hline
      Top-1 acc. (\%)    & 73.9                              & 78.1          & 80.2        & 82.1        & 83.1   & 83.9  \\
      \bottomrule
    \end{tabular}
  }
  \caption{ImageNet-1k training and evaluation settings for different PartialNet variants.}
  \label{tab:imagenet_settings}
\end{table}

\begin{table*}[ht]
  \centering
  \resizebox{1.0\linewidth}{!}{
    \begin{tabular}{@{}c|c|c|c|c|c|c|c|c|c@{}}
      \toprule
      Name                            & Output size                                & \multicolumn{2}{c|}{Layer specification}                                                                                                                                                                      & T0   & T1   & T2   & S     & M    & L           \\
      \hline
      Embedding                       & \large{$\frac{h}{4} \times \frac{w}{4}$}   & \begin{tabular}[c]{@{}c@{}}Conv\_4\_$c$\_4,\\ BN\end{tabular}                                                                                                                              & \# Channels $c$  & 32   & 48   & 64    & 96   & 128  & 160  \\
      \hline
      Stage 1                         & \large{$\frac{h}{4} \times \frac{w}{4}$}   & $\left[ \texttt{\begin{tabular}[c]{@{}c@{}}PAT\_ch\_3\_$c$\_1\_1/4,\\ Conv\_1\_$2c$\_1,\\ BN, Acti,\\ Conv\_1\_$c$\_1,\\ PAT\_sp\_1\_$c$\_1\_1/4 \end{tabular}}  \right] \times b_1 $      & \# Blocks $b_1$  & 1    & 1    & 2     & 2    & 2    & 2    \\
      \hline
      Merging                         & \large{$\frac{h}{8} \times \frac{w}{8}$}   & \begin{tabular}[c]{@{}c@{}}Conv\_2\_$2c$\_2,\\ BN\end{tabular}                                                                                                                             & \# Channels $2c$ & 64   & 96   & 128   & 192  & 256  & 320  \\
      \hline
      Stage 2                         & \large{$\frac{h}{8} \times \frac{w}{8}$}   & $\left[ \texttt{\begin{tabular}[c]{@{}c@{}}PAT\_ch\_3\_$2c$\_1\_1/4,\\ Conv\_1\_$4c$\_1,\\ BN, Acti,\\ Conv\_1\_$2c$\_1,\\ PAT\_sp\_1\_$2c$\_1\_1/4 \end{tabular}}  \right] \times b_2 $   & \# Blocks $b_2$  & 2    & 2    & 2     & 2    & 3    & 3    \\
      \hline
      Merging                         & \large{$\frac{h}{16} \times \frac{w}{16}$} & \begin{tabular}[c]{@{}c@{}}Conv\_2\_$4c$\_2,\\ BN\end{tabular}                                                                                                                             & \# Channels $4c$ & 128  & 192  & 256   & 384  & 512  & 640  \\
      \hline
      Stage 3                         & \large{$\frac{h}{16} \times \frac{w}{16}$} & $\left[ \texttt{\begin{tabular}[c]{@{}c@{}}PAT\_ch\_3\_$4c$\_1\_1/4,\\ Conv\_1\_$8c$\_1,\\ BN, Acti,\\ Conv\_1\_$4c$\_1,\\ PAT\_sp\_1\_$4c$\_1\_1/4 \end{tabular}}  \right] \times b_3 $   & \# Blocks $b_3$  & 8    & 8    & 6     & 9    & 16   & 20   \\
      \hline
      Merging                         & \large{$\frac{h}{32} \times \frac{w}{32}$} & \begin{tabular}[c]{@{}c@{}}Conv\_2\_$8c$\_2,\\ BN\end{tabular}                                                                                                                             & \# Channels $8c$ & 256  & 384  & 512   & 768  & 1024 & 1280 \\
      \hline
      Stage 4                         & \large{$\frac{h}{32} \times \frac{w}{32}$} & $\left[ \texttt{\begin{tabular}[c]{@{}c@{}}PAT\_ch\_3\_$8c$\_1\_1/4,\\ Conv\_1\_$16c$\_1,\\ BN, Acti,\\ Conv\_1\_$8c$\_1,\\ PAT\_sf\_1\_$8c$\_1\_1/4 \end{tabular}}  \right] \times b_4 $  & \# Blocks $b_4$  & 2    & 2    & 4     & 4    & 4    & 4    \\
      \hline
      Classifier                      & $1  \times 1$                              & \begin{tabular}[c]{@{}c@{}}Global average pool,\\ Conv\_1\_1280\_1,\\ Acti,\\ FC\_1000\end{tabular}                                                                                        & Acti             & GELU & GELU & ReLU  & ReLU & ReLU & ReLU \\
      \hline
      \multicolumn{4}{c|}{Params (M)} & 4.3                                        & 7.8                                                                                                                                                                                        & 12.6             & 29.0 & 61.3 & 104.4                      \\
      \hline
      \multicolumn{4}{c|}{FLOPs (G)}  & 0.25                                       & 0.55                                                                                                                                                                                       & 1.03             & 2.71 & 6.69 & 11.91                       \\
      \bottomrule
    \end{tabular}%
  }
  \caption{Configurations of different PartialNet variants. 'Conv\_$k$\_$c$\_$s$' refers to a convolutional layer with a kernel size of $k$, output channels of $c$, and a stride of $s$. 'PAT\_ch$\_k\_c\_s\_r$' refers to a partial attention convolution with an additional parameter, the split ratio $r$ of feature map channels, compared to a regular convolution. Similarly, 'PAT\_sp$\_k\_c\_s\_r$' and 'PAT\_sf$\_k\_c\_s\_r$' have the same configuration. Additionally, 'FC\_1000' refers to a fully connected layer with 1000 output channels. The $h \times w$ represents the input size, while $b\_i$ denotes the number of PartialNet blocks at stage $i$. FLOPs are calculated for an input size of $224 \times 224$.}
  \label{tab:configuration}
\end{table*}

Secondly, the configurations of different PartialNet variants are presented in~\cref{tab:configuration}. We also provide the detailed structures of the three different PATConv blocks, as shown in~\cref{fig:attention_block}.

\section{Full Comparison on the ImageNet-1k Benchmark and COCO Benchmark.}
For the full comparison of the classification task on the ImageNet-1k Benchmark, please refer to~\cref{tab:appendix_class_Benchmark}, which complements the results provided in Table~1 of the main paper. For the full Comparison of the object detection and instance segmentation tasks on the COCO2017 dataset please refer to~\cref{tab:appendix_dect_and_seg_Benchmark}, which complements the results provided in Table~2 of the main paper. 

\begin{table*}[ht] \small
  \centering
  \resizebox{1.0\linewidth}{!}{
  \begin{tabular}{@{}lcccccccc@{}}
    \toprule
    Network   &\makecell{Type}   & \makecell{Params\\(M)}	&\makecell{FLOPs\\(G)}	  &\makecell{Throughput\\V100 (FPS)$\uparrow$} &\makecell{Throughput\\MI250 (FPS)$\uparrow$}	&\makecell{Latency\\CPU (ms)$\downarrow$} &\makecell{Top-1\\(\%)$\uparrow$}  \\
    \hline
    ShuffleNetV2 x1.5\cite{Ma2018}      & CNN             & 3.5              & 0.30  & 5315       & 6642        & 13.7        & 72.6       \\
    MobileNetV2\cite{Sandler2018a}      & CNN             & 3.5              & 0.31  & 3924       & 7359        & 13.7        & 72.0       \\
    FasterNet-T0\cite{Chen2023}         & CNN             & 3.9              & 0.34  & {\bf 8546} & 10612       & {\bf 10.5}  & 71.9       \\
    MobileViT-XXS\cite{Mehta2021}       & Hybrid          & 1.3              & 0.42  & 2900       & 3321        & 16.7        & 69.0       \\
    MobileViTv2-0.5\cite{Mehta2022}     & Hybrid          & 1.4              & 0.46  & 3094       & 3135        & 15.8        & 70.2       \\
    PartialNet-T0(ours)                 & Hybrid          & 4.3              & 0.25  & 7777       & {\bf 11744} & 12.2        & {\bf 73.9} \\
    \hline
    EfficientNet-B0\cite{Tan2019}       & CNN             & 5.3              & 0.39  & 2934       & 3344        & 22.7        & 77.1       \\
    GhostNet x1.3\cite{Han2020}         & CNN             & 7.4              & 0.24  & 3788       & 3620        & 16.7        & 75.7       \\
    ShuffleNetV2 x2\cite{Ma2018}        & CNN             & 7.4              & 0.59  & 4290       & 5371        & 22.6        & 74.9       \\
    MobileNetV2 x1.4\cite{Sandler2018a} & CNN             & 6.1              & 0.60  & 2615       & 4142        & 21.7        & 74.7       \\
    FasterNet-T1\cite{Chen2023}         & CNN             & 7.6              & 0.85  & {\bf 4648} & 7198        & 22.2        & 76.2       \\
    EfficientViT-B1-192\cite{Cai2023b}  & Hybrid          & 9.1              & 0.38  & 4072       & 3912        & {\bf 19.3}  & 77.7       \\
    MobileViT-XS\cite{Mehta2021}        & Hybrid          & 2.3              & 1.05  & 1663       & 1884        & 32.8        & 74.8       \\
    PartialNet-T1(ours)                 & Hybrid          & 7.8              & 0.55  & 4403       & {\bf 7379}  & 21.5        & {\bf 78.1} \\
    \hline
    EfficientNet-B1\cite{Tan2019}       & CNN             & 7.8              & 0.70  & 1730       & 1583        & 35.5        & 79.1       \\
    ResNet50\cite{He2015}               & CNN             & 25.6             & 4.11  & 1258       & 3135        & 94.8        & 78.8       \\
    FasterNet-T2\cite{Chen2023}         & CNN             & 15.0             & 1.91  & 2455       & 4189        & 43.7        & 78.9       \\
    PoolFormer-S12\cite{Yu2022a}        & Hybrid          & 11.9             & 1.82  & 1927       & 3558        & 56.1        & 77.2       \\
    MobileViT-S\cite{Mehta2021}         & Hybrid          & 5.6              & 2.03  & 1219       & 1370        & 52.4        & 78.4       \\
    MobileViTv2-1.0\cite{Mehta2022}     & Hybrid          & 4.9              & 1.85  & 1391       & 1543        & 41.5        & 78.1       \\
    EfficientViT-B1\cite{Cai2023b}      & Hybrid          & 9.1              & 0.52  & 3072       & 3387        & {\bf 25.7}  & 79.4       \\
    PartialNet-T2(ours)                 & Hybrid          & 12.6             & 1.03  & {\bf 3074} & {\bf 4761}  & 35.2        & {\bf 80.2} \\
    \hline
    EfficientNet-B3\cite{Tan2019}       & CNN             & 12.0             & 1.80  & 768        & 926         & 73.5        & 81.6       \\
    ConvNeXt-T\cite{Liu2022b}           & CNN             & 28.6             & 4.47  & 902        & 1103        & 99.4        & {\bf 82.1} \\
    FasterNet-S\cite{Chen2023}          & CNN             & 31.1             & 4.56  & 1261       & 2243        & 96.0        & 81.3       \\
    PoolFormer-S36\cite{Yu2022a}        & Hybrid          & 30.9             & 5.00  & 675        & 1092        & 152.4       & 81.4       \\
    MobileViTv2-1.5\cite{Mehta2022}     & Hybrid          & 10.6             & 4.00  & 812        & 1000        & 104.4       & 80.4       \\
    MobileViTv2-2.0\cite{Mehta2022}     & Hybrid          & 18.5             & 7.50  & 551        & 684         & 103.7       & 81.2       \\
    Swin-T\cite{Liu2021c}               & Hybrid          & 28.3             & 4.51  & 808        & 1192        & 107.1       & 81.3       \\
    PartialNet-S(ours)                  & Hybrid          & 29.0             & 2.71  & {\bf 1559} & {\bf 2422}  & {\bf 72.5}  & {\bf 82.1} \\
    \hline
    EfficientNet-B4\cite{Tan2019}       & CNN             & 19.0             & 4.20  & 356        & 442         & 156.9       & 82.9       \\
    ConvNeXt-S\cite{Liu2022b}           & CNN             & 50.2             & 8.71  & 510        & 610         & 185.5       & {\bf 83.1} \\
    FasterNet-M\cite{Chen2023}          & CNN             & 53.5             & 8.74  & 621        & 1098        & 181.6       & 83.0       \\
    PoolFormer-M36\cite{Yu2022a}        & Hybrid          & 56.2             & 8.80  & 444        & 721         & 244.3       & 82.1       \\
    Swin-S\cite{Liu2021c}               & Hybrid          & 49.6             & 8.77  & 477        & 732         & 199.1       & 83.0       \\
    PartialNet-M(ours)                  & Hybrid          & 61.3             & 6.69  & {\bf 799}  & {\bf 1280}  & {\bf 155.3} & {\bf 83.1} \\
    \hline
    EfficientNet-B5\cite{Tan2019}       & CNN             & 30.0             & 9.90  & 246        & 313         & 333.3       & 83.6       \\
    ConvNeXt-B\cite{Liu2022b}           & CNN             & 88.6             & 15.38 & 322        & 430         & 317.1       & 83.8       \\
    FasterNet-L\cite{Chen2023}          & CNN             & 93.5             & 15.52 & 384        & 709         & 312.5       & 83.5       \\
    PoolFormer-M48\cite{Yu2022a}        & Hybrid          & 73.5             & 11.59 & 335        & 556         & 322.3       & 82.5       \\
    Swin-B\cite{Liu2021c}               & Hybrid          & 87.8             & 15.47 & 315        & 520         & 333.8       & 83.5       \\
    PartialNet-L(ours)                  & Hybrid          & 104.3            & 11.91 & {\bf 426}  & {\bf 765}   & {\bf 272.5} & {\bf 83.9} \\
    \bottomrule
  \end{tabular}
  }
  \caption{Full comparison on ImageNet-1k Benchmark: models with similar top-1 accuracy are grouped together. The best results are in bold.}
  \label{tab:appendix_class_Benchmark}
\end{table*}

\begin{table*}[ht] \small
  \centering
  \resizebox{1.0\linewidth}{!}{
  \begin{tabular}{@{}lccccc cccc@{}}
    \toprule
    Backbone              & \makecell{Params\\(M)}	&\makecell{FLOPs\\(G)} &\makecell{Throughput\\MI250 (FPS)$\uparrow$}	&\makecell{$AP^{b}\uparrow$}  &\makecell{$AP^{b}_{50}\uparrow$} &\makecell{$AP^{b}_{75}\uparrow$} &\makecell{$AP^{m}\uparrow$} &\makecell{$AP^{m}_{50}\uparrow$} &\makecell{$AP^{m}_{75}\uparrow$}\\
    \hline
    ResNet50\cite{He2015}                 & 44.2    & 253   &121         & 38.0       & 58.6       & 41.4       & 34.4       & 55.1       & 36.7       \\
    PoolFormer-S24\cite{Yu2022a}          & 41.0    & 233   &68          & 40.1       & 62.2       & 43.4       & 37.0       & 59.1       & 39.6       \\
    PVT-Small x1.5\cite{Wang2021c}        & 44.1    & 238   &98          & 40.4       & 62.9       & 43.8       & 37.8       & 60.1       & 40.3       \\
    FasterNet-S\cite{Chen2023}            & 49.0    & 258   &121         & 39.9       & 61.2       & 43.6       & 36.9       & 58.1       & 39.7       \\
    PartialNet-S(ours)                    & 46.9    & 216   &{\bf 122}   & {\bf 42.7} & {\bf 64.9} & {\bf 46.5} & {\bf 39.3} & {\bf 61.8} & {\bf 42.2} \\
    \hline
    ResNet101\cite{Mehta2021}             & 63.2    & 329   &62          & 40.4       & 61.1       & 44.2       & 36.4       & 57.7       & 38.8       \\
    ResNeXt101-32$\times$4d\cite{Xie2017} & 62.8    & 333   &51          & 41.9       & 62.5       & 45.9       & 37.5       & 59.4       & 40.2       \\
    PoolFormer-S36\cite{Yu2022a}          & 50.5    & 266   &44          & 41.0       & 63.1       & 44.8       & 37.7       & 60.1       & 40.0       \\
    PVT-Medium\cite{Wang2021c}            & 63.9    & 295   &52          & 42.0       & 64.4       & 45.6       & 39.0       & 61.6       & 42.1       \\
    FasterNet-M\cite{Chen2023}            & 71.2    & 344   &62          & 43.0       & 64.4       & 47.4       & 39.1       & 61.5       & 42.3       \\
    PartialNet-M(ours)                    & 78.2    & 295   &{\bf 65}    & {\bf 44.3} & {\bf 65.8} & {\bf 48.5} & {\bf 40.6} & {\bf 63.3} & {\bf 43.7} \\
    \hline
    ResNeXt101-64$\times$4d\cite{Xie2017} & 101.9   & 487   &29          & 42.8       & 63.8       & 47.3       & 38.4       & 60.6       & 41.3       \\
    PVT-Large$\times$4d\cite{Wang2021c}   & 81.0    & 358   &26          & 42.9       & 65.0       & 46.6       & 39.5       & 61.9       & 42.5       \\
    FasterNet-L\cite{Chen2023}            & 110.9   & 484   &35          & 44.0       & 65.6       & 48.2       & 39.9       & 62.3       & 43.0       \\
    PartialNet-L(ours)                    & 122.0   & 397   &{\bf 39}    & {\bf 44.7} & {\bf 66.3} & {\bf 49.0} & {\bf 41.0} & {\bf 63.7} & {\bf 44.2} \\
    \bottomrule
  \end{tabular}
  }
  \caption{Results using PartialNet-S/M/L on object detection and instance segmentation benchmark in COCO dataset.}
  \label{tab:appendix_dect_and_seg_Benchmark}
\end{table*}

\section{The Complements of Ablation Studies}
\textbf{ Partial Attention vs. Classic Visual Attention:}
To further prove the superiority of our proposed PATConv, we present experiment results for the combination of our partial attention and classic visual attention networks, and the results are shown in \cref{tab:abla_convention_attention_acc}. The results demonstrate the effectiveness of our enhanced PATConv block, .e., PAT\_ch.

\textbf{Partial Attention Convolution (PATConv) vs. Partial Convolution (PConv) Under the Same Training Settings.}
In order to verify the effectiveness and ensure a fair comparison of our PartialNet, we reproduced the results of all FastNet variants on ImageNet-1k using our training experiment configuration. The results are shown in~\cref{tab:pat_vs_fasternet}. It can be seen from the results that our PartialNet still has significant advantages.

\begin{table*}[ht]
  \centering
  \resizebox{0.8\linewidth}{!}{
  \begin{tabular}{@{}lcccc c@{}}
    \toprule
    Visual type & \makecell{Params(M)} & \makecell{FLOPs(G)} & \makecell{Throughput(fps)$\uparrow$} & \makecell{latency(ms)$\downarrow$} & \makecell{Acc1(\%)$\uparrow$} \\
    \hline
    SRM~\cite{Lee2019a}               & 12.2       & 1.03     & 4751         & 35.2        & 79.6                          \\
    SE-NET~\cite{Hu2018}              & 12.3       & 1.04     & 4910         & 32.3        & 79.8                          \\
    PAT(ours)                         & 12.6       & 1.03     & 4761         & 35.2        & {\bf 80.2}                     \\
    \bottomrule
  \end{tabular}
  }
  \caption{Comparison on PartialNet-T2 of partial visual attention and conventional visual attention on ImageNet1K dataset.}
  \label{tab:abla_convention_attention_acc}
\end{table*}

\begin{table*}[ht] \small
  \centering
  \resizebox{0.87\linewidth}{!}{
  \begin{tabular}{@{}lcccccccc@{}}
    \toprule
    Network   &\makecell{Type}   & \makecell{Params\\(M)}	&\makecell{FLOPs\\(G)}	  &\makecell{Throughput\\V100 (FPS)$\uparrow$} &\makecell{Throughput\\MI250 (FPS)$\uparrow$}	&\makecell{Latency\\CPU (ms)$\downarrow$} &\makecell{Top-1\\(\%)$\uparrow$}  \\
    \hline
    FasterNet-T0\cite{Chen2023}         & CNN             & 3.9              & 0.34  & {\bf 8546} & 10612       & {\bf 10.5}  & 71.9       \\
    FasterNet-T0*\cite{Chen2023}        & CNN             & 3.9              & 0.34  & {\bf 8546} & 10612       & {\bf 10.5}  & 71.0       \\
    PartialNet-T0(ours)                 & Hybrid          & 4.3              & 0.25  & 7777       & {\bf 11744} & 12.2        & {\bf 73.9} \\
    \hline
    FasterNet-T1\cite{Chen2023}         & CNN             & 7.6              & 0.85  & {\bf 4648} & 7198        & 22.2        & 76.2       \\
    FasterNet-T1*\cite{Chen2023}        & CNN             & 7.6              & 0.85  & {\bf 4648} & 7198        & 22.2        & 76.5       \\
    PartialNet-T1(ours)                 & Hybrid          & 7.8              & 0.55  & 4403       & {\bf 7379}  & 21.5        & {\bf 78.1} \\
    \hline
    FasterNet-T2\cite{Chen2023}         & CNN             & 15.0             & 1.91  & 2455       & 4189        & 43.7        & 78.9       \\
    FasterNet-T2*\cite{Chen2023}        & CNN             & 15.0             & 1.91  & 2455       & 4189        & 43.7        & 79.2       \\
    PartialNet-T2(ours)                 & Hybrid          & 12.6             & 1.03  & {\bf 3074} & {\bf 4761}  & 35.2        & {\bf 80.2} \\
    \hline
    FasterNet-S\cite{Chen2023}          & CNN             & 31.1             & 4.56  & 1261       & 2243        & 96.0        & 81.3       \\
    FasterNet-S\cite{Chen2023}          & CNN             & 31.1             & 4.56  & 1261       & 2243        & 96.0        & 81.5       \\
    PartialNet-S(ours)                  & Hybrid          & 29.0             & 2.71  & {\bf 1559} & {\bf 2422}  & {\bf 72.5}  & {\bf 82.1} \\
    \hline
    FasterNet-M\cite{Chen2023}          & CNN             & 53.5             & 8.74  & 621        & 1098        & 181.6       & 83.0       \\
    FasterNet-M*\cite{Chen2023}         & CNN             & 53.5             & 8.74  & 621        & 1098        & 181.6       & 83.0       \\
    PartialNet-M(ours)                  & Hybrid          & 61.3             & 6.69  & {\bf 799}  & {\bf 1280}  & {\bf 155.3} & {\bf 83.1} \\
    \hline
    FasterNet-L\cite{Chen2023}          & CNN             & 93.5             & 15.52 & 384        & 709         & 312.5       & 83.5       \\
    FasterNet-L*\cite{Chen2023}         & CNN             & 93.5             & 15.52 & 384        & 709         & 312.5       & 83.6       \\
    PartialNet-L(ours)                  & Hybrid          & 104.3            & 11.91 & {\bf 426}  & {\bf 765}   & {\bf 272.5} & {\bf 83.9} \\
    \bottomrule
  \end{tabular}
  }
  \caption{Comparison on ImageNet-1k. The "*" denotes reproduced results based on our experimental setups.}
  \label{tab:pat_vs_fasternet}
\end{table*}

\end{document}